\documentclass[10pt,twocolumn,letterpaper]{article}

\usepackage{wacv}
\usepackage{times}
\usepackage{epsfig}
\usepackage{graphicx}
\usepackage{amsmath}
\usepackage{amssymb}
\usepackage{booktabs}
\usepackage{multirow}
\usepackage{multicol}
\usepackage{authblk}
\usepackage[accsupp]{axessibility}  %

\def\wacvPaperID{1686} %

\wacvalgorithmstrack   %
\wacvfinalcopy %

\ifwacvfinal
\usepackage[breaklinks=true,bookmarks=false]{hyperref}
\else
\usepackage[pagebackref=true,breaklinks=true,colorlinks,bookmarks=false]{hyperref}
\fi

\begin{document}

\title{Kinematic-aware Hierarchical Attention Network\\ for Human Pose Estimation in Videos}

\author[1]{Kyung-Min Jin}
\author[1]{Byoung-Sung Lim}
\author[2]{Gun-Hee Lee}
\author[1]{Tae-Kyung Kang}
\author[1]{Seong-Whan Lee}
\affil[1]{Department of Artificial Intelligence, Korea University}
\affil[2]{Department of Computer Science and Engineering, Korea University}
\affil[ ]{\tt\small \{km\_jin, bs\_lim, gunhlee, tk\_kang, sw.lee\}@korea.ac.kr}

\maketitle
\thispagestyle{empty}

\begin{abstract}
     Previous video-based human pose estimation methods have shown promising results by leveraging aggregated features of consecutive frames. However, most approaches compromise accuracy to mitigate jitter or do not sufficiently comprehend the temporal aspects of human motion. Furthermore, occlusion increases uncertainty between consecutive frames, which results in unsmooth results. To address these issues, we design an architecture that exploits the keypoint kinematic features with the following components. First, we effectively capture the temporal features by leveraging individual keypoint's velocity and acceleration. Second, the proposed hierarchical transformer encoder aggregates spatio-temporal dependencies and refines the 2D or 3D input pose estimated from existing estimators. Finally, we provide an online cross-supervision between the refined input pose generated from the encoder and the final pose from our decoder to enable joint optimization. We demonstrate comprehensive results and validate the effectiveness of our model in various tasks: 2D pose estimation, 3D pose estimation, body mesh recovery, and sparsely annotated multi-human pose estimation. Our code is available at \url{https://github.com/KyungMinJin/HANet}.
\end{abstract}
\section{Introduction}

\begin{figure}[t]
\centering
\includegraphics[width=1.0\linewidth]{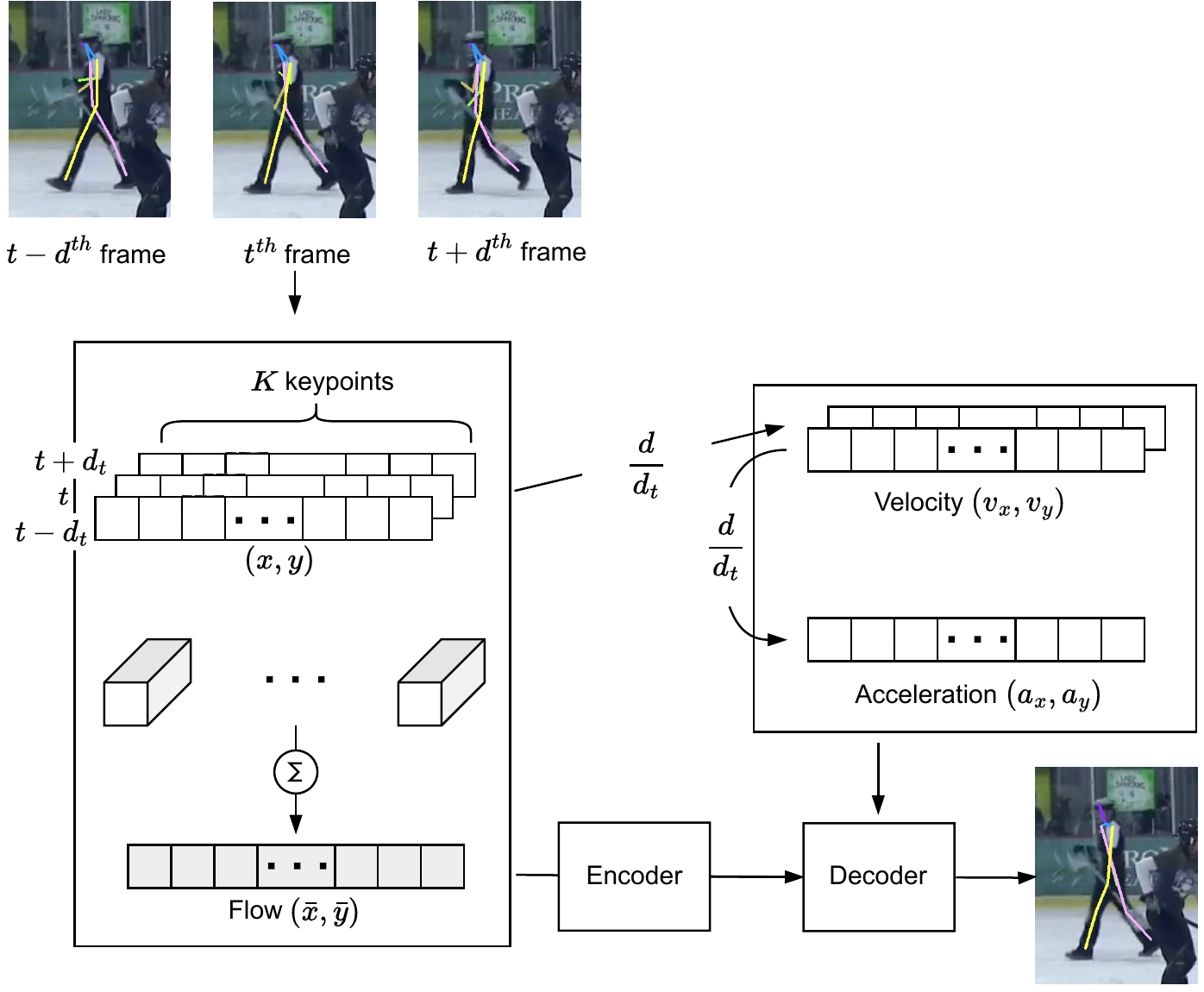}
\caption{The overview of HANet with keypoint kinematic features. First, we compute flow as a weighted sum of keypoint coordinates from current, previous, and next frames and hierarchically encode the temporal features of body motion. In addition, we further consider velocity $v$ and acceleration $a$ from consecutive keypoint coordinates as supplement input of the decoder.}
\label{fig:keypoint motion}
\end{figure}

Human pose estimation, estimating each keypoint location from images, has long been studied in the computer vision field. It has been extended from identifying a person's location to action understanding \cite{actrecog_vi, htnet, tanfous2022and, ahmad2006human, roh2010view, roh2007accurate} or manipulating various computer interfaces with human movements, thereby becoming a core technology for various applications. In addition, with the advent of deep learning, it became possible to locate each keypoint robustly, and methods \cite{hrnet,fastpose,partaffinityfields,alphapose,alpha++,yang2007reconstruction,lee2021uncertainty} have shown remarkable results. However, existing methods still fail to address highly-occluded cases, such as the presence of multiple persons or motion blur where certain body parts move quickly. Thus, their results include high-frequency jitters.

We found two significant issues in pose estimation for video. First, the large positional difference between consecutive frames' poses shares less visual similarity, resulting in high-frequency jitters. Second, occluded or blurred regions bring spatial ambiguities that significantly drop model performance, and make the task more challenging. 

Existing methods \cite{lpm,3dspatiotemporal,deciwatch,transfusion,skeletor,dkd} tend to focus on one of the issues instead of both; reducing jitter by focusing on the temporal aspect or addressing occlusion by increasing the model complexity to capture spatial features well. In video-based pose estimation, existing methods typically use recurrent neural networks (RNN) \cite{lpm}, 3D convolutional neural networks (3D CNN) \cite{3dhrnet}, and transformers \cite{3dspatiotemporal,deciwatch,transfusion,skeletor} to exploit temporal features. However, they show limitations where an input video includes severe occlusion or motion blur. Although there are methods \cite{dcpose,posewarper,3dhrnet,dkd,margipose} using 2D CNN to store spatio-temporal dependencies within parameters, they do not accurately comprehend the temporal features of human motion.

In this paper, we propose a novel architecture named \textbf{HANet} (Kinematic-aware Hierarchical Attention Network) that accurately refines human poses in video. We introduce a hierarchical network that effectively addresses jitter and occlusion by exploiting keypoints' movement, as shown in Fig.~\ref{fig:keypoint motion}. First, we compute each keypoint's kinematic features: flow (the track of keypoint movement), velocity, and acceleration. Through these features, our framework kinematically learns the temporal aspect of keypoints to focus on frequently occluded or fast-moving body parts such as wrists and ankles. Second, the proposed hierarchical encoder projects multi-scale feature maps by exponentially increasing the number of channels and captures spatio-temporal features. We embed multi-scale feature maps to generate positional offsets which we add to refine input poses, estimated by off-the-shelf methods. Then, our decoder processes the refined input poses with keypoint velocity and acceleration to estimate the final poses. Lastly, we provide a cross-supervision that cooperatively optimizes refined input poses and final poses by choosing an online learning target along their training losses. Through this work, our method significantly reduces jitter and becomes robust to occlusion while improving performance. In summary, our main contributions are as follows:
\begin{itemize}
    \item We propose a novel approach HANet using keypoint kinematic features, following the laws of physics. Our method effectively mitigates jitter and becomes robust to occlusion with these proposed features. 
    \item We propose a hierarchical transformer encoder that incorporates multi-scale spatio-temporal attention. We leverage all layers' multi-scale attention maps and improve performance on sparse supervision.
    \item We propose online mutual learning that enables joint optimization of refined input poses and final poses by selecting online targets along their training losses. 
    \item We conduct extensive experiments and demonstrate our framework's applicability on various tasks: 2D and 3D pose estimation, body mesh recovery, and sparsely-annotated multi-human 2D pose estimation.
\end{itemize}
\section{Related Work}

\subsection{Pose Estimation in Images}

Modern approaches for single-image pose estimation, one of the fundamental pattern recognition problems \cite{fujisawa1999information,lee1990translation,lee1999integrated,pr7} in computer vision field, are typically based on 2D CNN. Early methods \cite{toshev2014deeppose} directly regress the joint coordinates from the images; however, recent approaches \cite{hrnet,partaffinityfields,cpm,fcpose,hourglass} have widely adopted joint position representations with maximum values from a heatmap that denotes joint presence probability. These methods can be divided into two ways: bottom-up and top-down. First, viewing a human skeleton as a graph, bottom-up approaches \cite{partaffinityfields,deepcut} detect individual body parts and assemble these components into the person. Recently, many top-down methods \cite{hourglass,cpm,hrnet} have been proposed that first perform human detection using an object detector and estimate poses for each individual. CPM \cite{cpm} iteratively refines the output of each step, and Hourglass \cite{hourglass} adjusts the number of channels. HRNet \cite{hrnet} has achieved higher performance than ResNet \cite{resnet} by maintaining high-resolution feature maps through multi-scale fusion and replaced ResNet, which served as the backbone of pose estimation. These top-down methods have significantly improved performance and show remarkable results, but they include high-frequency jitter when applied to video data.

\begin{figure*}[t]
\centering
\includegraphics[width=1.0\textwidth]{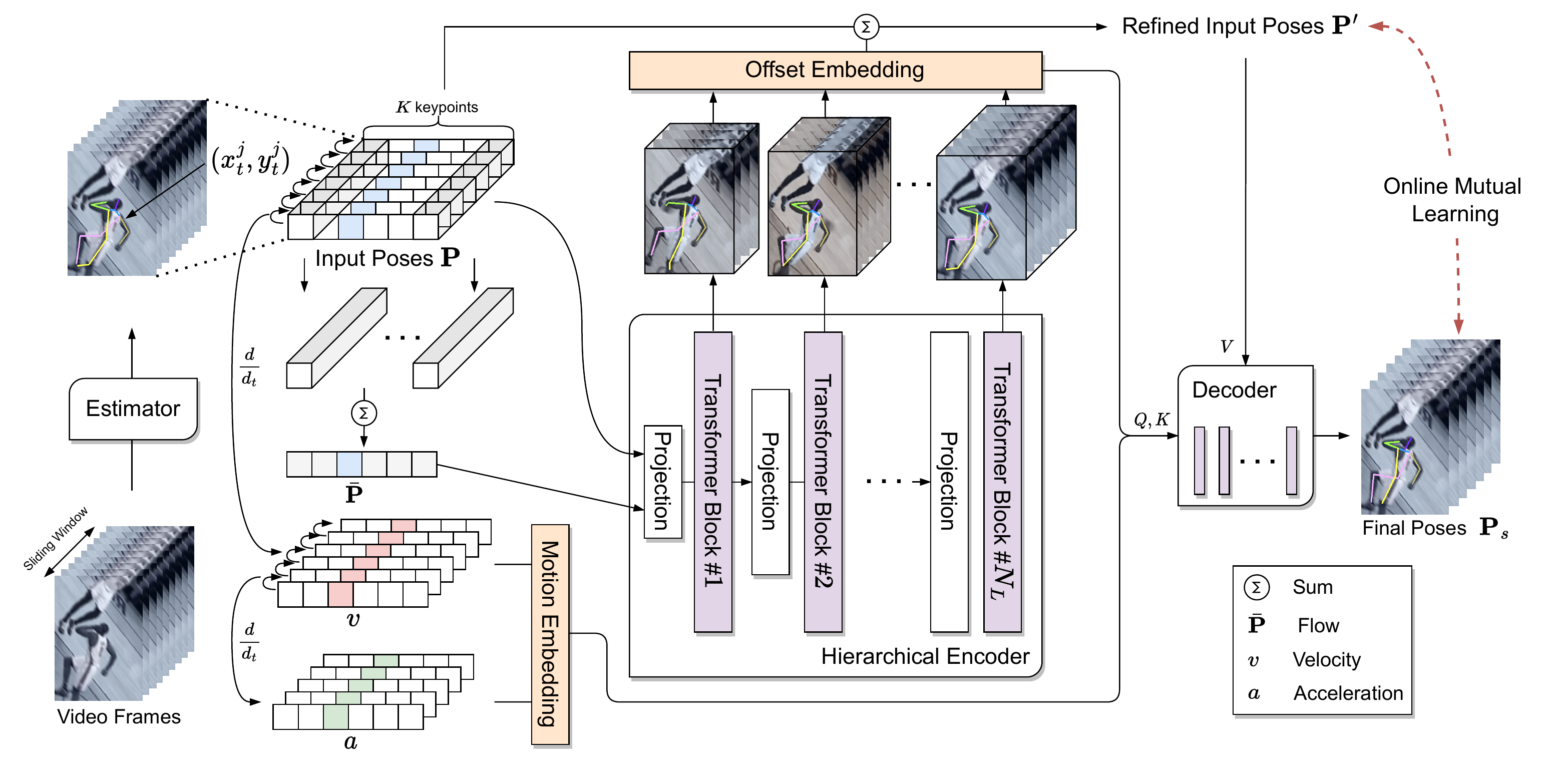}
\caption{The overall architecture of HANet. First, keypoint kinematic features (flow, velocity, and acceleration) are computed from input poses. Then, our encoder captures the spatio-temporal relationships of keypoint movement. Each encoder layer features are embedded to offsets which refine input poses. Refined input poses are decoded with keypoint velocity and acceleration to produce final poses.}
\label{fig:pipeline}
\end{figure*}

With the success of the attention-based method in natural language processing (NLP), transformers combined with CNNs have brought inspiration for new approaches \cite{vivit, vit, transpose, 3dspatiotemporal, skeletor,transfusion,epipolar,tokenpose,otpose} to computer vision field. Thanks to the characteristic of self-attention, which shows superior performance in modeling long-range dependencies, the transformer can be used to capture spatio-temporal relations. With the advent of ViT \cite{vit}, which outperformed CNN-based counterparts in classification on large image datasets \cite{coco}, several methods \cite{tokenpose,transpose} applied transformers to pose estimation. Transpose \cite{transpose} captures long-range relationships and reveals spatial dependencies that provide evidence of how the model handles occlusion. Tokenpose \cite{tokenpose} tokenizes each keypoint and computes an attention map between keypoints. However, they increased the model size and the resolution of input images (or heatmaps), making it difficult to apply transformer for videos with many frames.

\subsection{Pose Estimation in Videos}

Fully-annotated benchmark datasets \cite{jhmdb,h36m,3dpw,aist} for videos are suitable for learning temporal features because they provide supervision for all frames and contain a few people with less occlusion. Meanwhile, \cite{3dhrnet,deciwatch,lpm,dkd,kfp} directly process 2D or 3D positions and capture temporal features of body motion using RNN or 3D CNN. LPM \cite{lpm} extends CPM \cite{cpm} using long short-term memory (LSTM) to capture temporal dependencies between poses. 3D HRNet \cite{3dhrnet} intuitively uses temporal convolutions and learns the correlation between consecutive frames' keypoints by extending HRNet \cite{hrnet}. Recently, several approaches \cite{3dspatiotemporal, transfusion,skeletor,deciwatch} have also used vanilla transformers \cite{attentionisallyouneed} for pose estimation in videos. \cite{3dspatiotemporal} encodes 3D information using a spatial and temporal encoder. Also, \cite{transfusion,epipolar} use transformer to fuse multi-view features. Recently, DeciWatch \cite{deciwatch} proposes a method that efficiently watches sparsely sampled frames with transformer taking advantage of the continuity of human motions without performance degradation. However, these methods may not be useful in many real-world scenarios, as they do not perform well in crowded scenes with severe occlusion. 

Many CNN-based approaches \cite{poseflow,3dhrnet} have been proposed to address the occlusion issues in videos. The basic idea of CNN-based pose estimation methods for video is to encode spatio-temporal relationships between frames using convolution. However, there is a fundamental problem in that the receptive field is limited, which makes it challenging to capture long-range spatio-temporal relationships between joint positions. In addition, recent methods leverage memory-intensive heatmap-based estimation processes, making it challenging to sufficiently consider the temporal aspect of human motion. In contrast, we directly use input poses from estimators \cite{simplebaseline,dcpose,pare,fcn,spin,li2022mhformer} to train our framework HANet, which efficiently reduces memory usage and capturing the spatio-temporal dependencies by proposed hierarchical transformer encoder.

\section{Method}

We propose a novel framework that leverages keypoint kinematic features among consecutive frames to reduce jitter and learn temporal features of body motion. First, we use input poses $\mathbf{P}\in\mathbb{R}^{T/N\times(K\cdot D)}$ estimated from the off-the-shelf estimators \cite{dcpose, simplebaseline,fcn,pare,spin,li2022mhformer} to compute keypoint kinematic features within a sliding window of length $T$. Here, $K$ is the number of keypoints, $D$ denotes the dimension of each keypoint coordinates, and $N$ is the interval \cite{deciwatch} that we sampled poses in a sliding window. Then, we construct hierarchical transformer architecture that processes consecutive poses. The hierarchical encoder increases the number of channels exponentially and projects multi-scale feature maps to positional offsets that refine input poses $\mathbf{P}$. Finally, our decoder processes the offsets, keypoint velocity, and acceleration as keys and queries with refined input poses as values to estimate final poses. We discuss each component in more detail below.

\subsection{Keypoint Kinematic Features} 

We consider keypoint kinematic features within a sliding window to address jitter, a fundamental problem of pose estimation in videos. In this paper, we leverage three kinematic features, a continual aspect of human motion, obtained from previous, current, and next frames' poses.

\textbf{Keypoint Flow.} First, we compute the track of keypoint movement using consecutive poses' coordinates $\mathbf{P}_t$ at each frame $t\in\{1,2,\ldots,T\}$. 
We denote the keypoint flow as $\bar{\mathbf{P}}_t$ and define it as,

\begin{equation}
    \begin{aligned}
        \bar{\mathbf{P}}_t & = (\mathbf{P}_t + \mathbf{P}_{t + d_t} + \mathbf{P}_{t - d_t})/3,
    \end{aligned}
\end{equation}
where $d_t$ denotes an interval from the previous and next poses to the current poses. A flow $\bar{\mathbf{P}}_t$ can be interpreted as a moving average or a track of keypoint movement. 

\textbf{Keypoint Velocity and Acceleration.} We further consider kinematic information from the perspective of keypoint velocity and acceleration between consecutive input poses. We define the keypoint velocity and acceleration as below:
\begin{equation}
    \begin{aligned}
        &\boldsymbol{v}_{t} = (\mathbf{P}_t - \mathbf{P}_{t- d_t})/d_t,\\
        &\boldsymbol{a}_{t} = (\boldsymbol{v}_{t} - \boldsymbol{v}_{t -d_t})/d_t.
    \end{aligned}
\end{equation}
We exploit them in our decoder as first and second-order derivative features to estimate final poses.

\begin{figure}[t]
\centering
\includegraphics[width=1.0\linewidth]{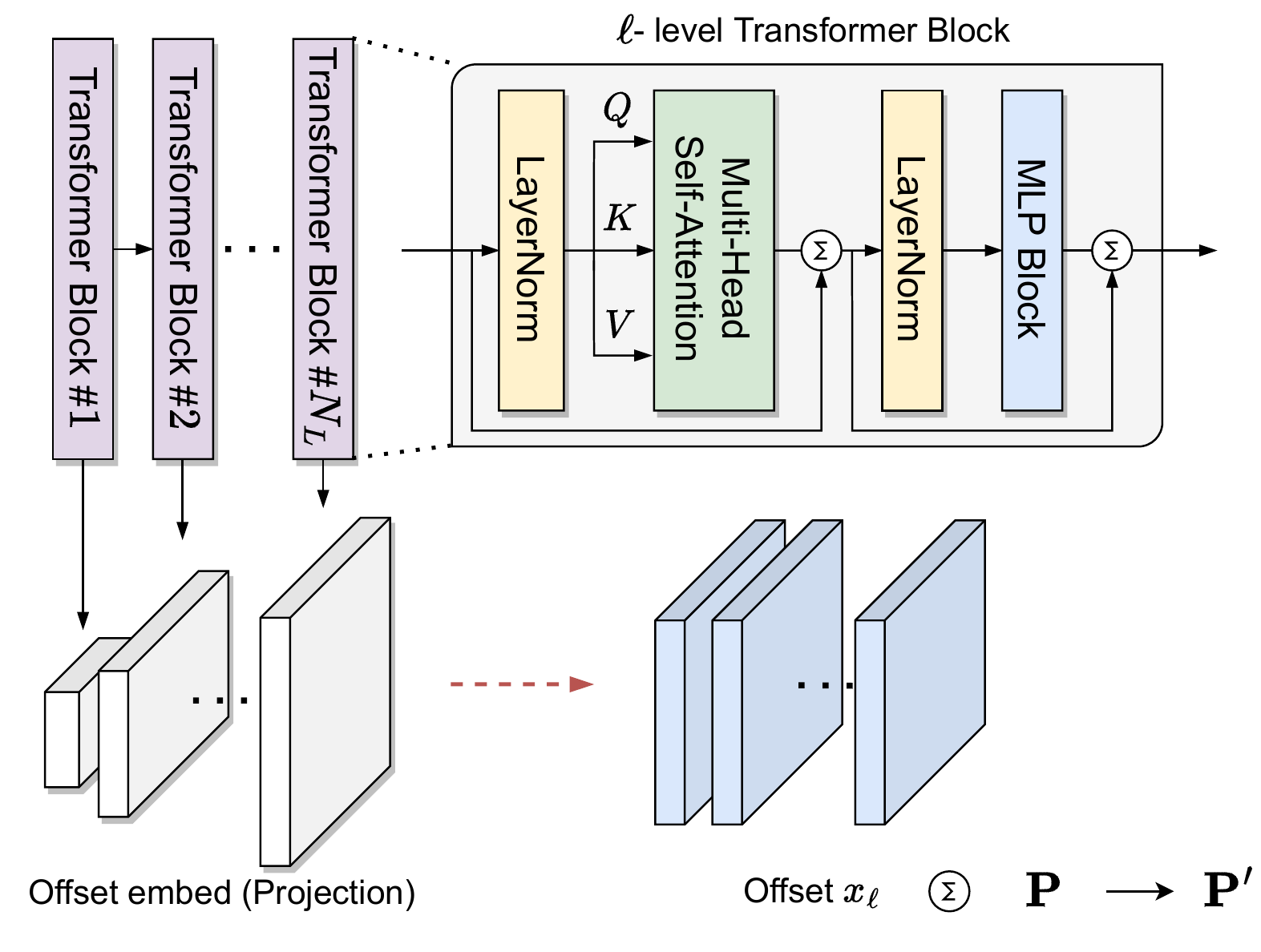}
\caption{Visualization of the proposed hierarchical encoder. Each encoder layer iteratively captures a spatio-temporal correlation between keypoints. Then attention map of each layer is linearly projected to produce positional offsets $\boldsymbol{x}_\ell$. We can get refined input poses $\mathbf{P}^\prime$ by a weighted sum of offsets $\mathbf{X}$ and input poses $\mathbf{P}$.}
\label{fig:hierarchical encoder}
\end{figure}

\subsection{Hierarchical Encoder}

After keypoint kinematic features are computed, the proposed hierarchical transformer encoder generates multi-scale pose features $\hat{\mathbf{Z}}_\ell\in\mathbb{R}^{T/N\times (C\cdot 2^{\ell})}$ at each layer $\ell$, representing spatio-temporal attention maps. Here, $C$ denotes the initial embedding dimension. Long-range and short-range attention maps model the distribution of large and small body movements, respectively, and resolve occlusion of keypoints spatially within a frame or temporally specific frame in a video.

First, we embed previous, current, next poses, and keypoint flow $\bar{\mathbf{P}}_t$ by stacking them in keypoint dimension. This can be defined as:
\begin{equation}
    \begin{aligned}
        &\mathbf{Z}_0 = (\bar{\mathbf{P}}_t;\mathbf{P}_{t}; \mathbf{P}_{t + d_t}; \mathbf{P}_{t - d_t})\mathbf{W}_{0},\\
    \end{aligned}
\end{equation}
where $\mathbf{W}_{0}\in\mathbb{R}^{(K\cdot 4D)\times C}$ is a projection matrix and $\mathbf{Z}_0$ denotes the initial embedding features. $\mathbf{Z}_0$ is then projected to $\mathbf{Q}_0$, $\mathbf{K}_0$, and $\mathbf{V}_0$ representing queries, keys, and values. The hierarchical encoder exponentially expands queries, keys, and values using $\mathbf{W}_{\ell}\in\mathbb{R}^{C\cdot 2^{\ell-1}\times C\cdot 2^{\ell}}$. These are given by,
\begin{equation}
    \begin{aligned}
        &\mathbf{Q}_{\boldsymbol\ell} = (\mathbf{Z}_{\ell-1})\mathbf{W}_{\ell} + \mathbf{E}_{pos,\ell}, &\boldsymbol\ell=1 \ldots N_L,\\
        &\mathbf{K}_{\boldsymbol\ell} = (\mathbf{Z}_{\ell-1})\mathbf{W}_{\ell} + \mathbf{E}_{pos,\ell}, &\boldsymbol\ell=1 \ldots N_L,\\
        &\mathbf{V}_{\boldsymbol\ell} = (\mathbf{Z}_{\ell-1})\mathbf{W}_{\ell} + \mathbf{E}_{pos,\ell}, &\boldsymbol\ell=1 \ldots N_L,\\
    \end{aligned}
\end{equation}
where $\mathbf{E}_{pos,\ell}\in\mathbb{R}^{T/N\times (C \cdot 2^\ell)}$ is an $\ell$-level positional embedding and $N_L$ denotes the number of encoder layers. Then, the hierarchical encoder generates a tuple of multi-scale feature maps $(\mathbf{Z}_0, \mathbf{Z}_1, \ldots , \mathbf{Z}_{N_L})$. We compute attention maps as a vanilla transformer \cite{attentionisallyouneed} and apply LayerNorm (LN) \cite{layernorm} before every multi-head self-attention (MSA) and multi-layer perceptron (MLP) block. Leaky ReLU \cite{lrelu} is used for the MLP activation function. If we simply express projected $\ell$-level queries, keys, and values as $\mathbf{Z}_\ell$, this process can be expressed as:
\begin{equation}
    \begin{aligned}
        &\overline{\mathbf{Z}}_{\ell}=\operatorname{MSA}(\operatorname{LN}(\mathbf{Z}_{\ell}))+\mathbf{Z}_{\ell},  \\
        &\hat{\mathbf{Z}}_{\ell}=\operatorname{MLP}(\operatorname{LN}(\overline{\mathbf{Z}}_{\ell}))+\overline{\mathbf{Z}}_{\ell},
    \end{aligned}
\end{equation}
where $\hat{\mathbf{Z}}_{\ell}$ is the encoded multi-scale spatio-temporal features. Then, we project the encoded multi-scale features to offset embeddings that coincide with the dimension of input poses $K\cdot D$, using 1D convolution $\operatorname{\textbf{Conv1D}}_\ell$. We denote the projected positional offsets as $\boldsymbol{x}_\ell$, as illustrated in Fig.~\ref{fig:hierarchical encoder}. A weighted sum of $\boldsymbol{x}_\ell$ is added to the input pose $\mathbf{P}$ and can be defined as follows:
\begin{equation}
    \begin{aligned}
        &\boldsymbol{x}_\ell = \operatorname{\textbf{Conv1D}}_\ell(\hat{\mathbf{Z}}_{\ell}),\\
        & \mathbf{P}^\prime = \mathbf{P} + \frac{1}{N_L} \sum_{\ell=0}^{N_L} \boldsymbol{x}_\ell,
    \end{aligned}
\end{equation}
where $\mathbf{P}^\prime$ denotes the refined input poses.

\begin{figure*}[t]
\centering
\includegraphics[width=1.0\linewidth]{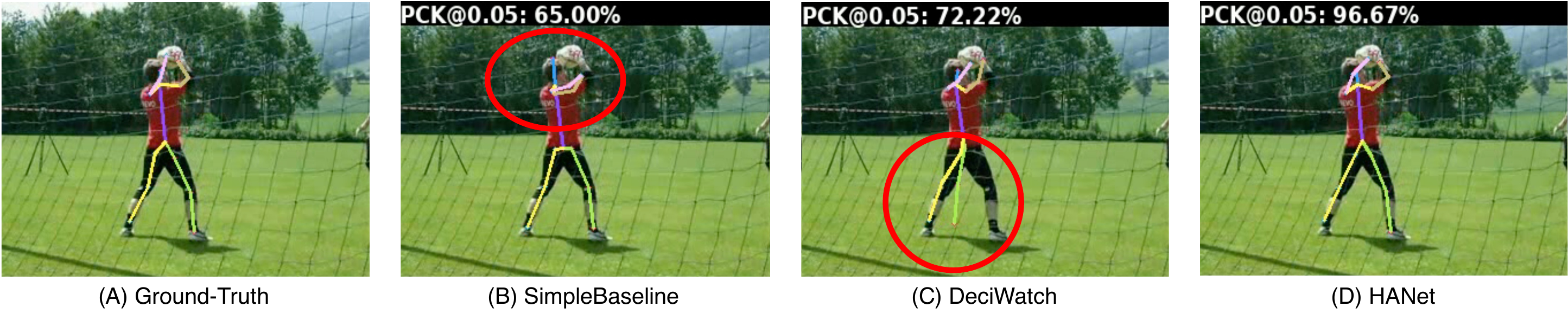}
\caption{Qualitative comparison on Sub-JHMDB \cite{jhmdb} dataset. From left to right, A, B, C, and D are ground truth, output of SimpleBaseline \cite{simplebaseline},  DeciWatch \cite{deciwatch}, and HANet. We report PCK@0.05 on the video and visualize that our framework outperforms existing methods. 
}
\label{fig:jhmdb qualitative}
\end{figure*}

\subsection{Decoder}
Given the weighted sum of offsets $\mathbf{X}$, the decoder processes them with $\boldsymbol{v}_t$ and $\boldsymbol{a}_t$ as keys and queries with $\mathbf{P}^\prime$ as values, to estimate the final poses $\textbf{P}_s$. First, we use $\boldsymbol{v}_t$ and $\boldsymbol{a}_t$ as derivative features with 1D convolutional layers, which can be defined as,
\begin{equation}
    \begin{aligned}
        &\textbf{S}_v = w_v \boldsymbol{v}_t + b_v,\\
        &\textbf{S}_a = w_a \boldsymbol{a}_t^2 + b_a,
     \end{aligned}
\end{equation}
to represent the derivative positional features $\textbf{S}_v$ and $\textbf{S}_a$, where $w_v, w_a\in\mathbb{R}^{C\times C}$ and $b_v, b_a\in\mathbb{R}^C$ are weights and biases. Then, we stack them along keypoint channels and leverage the transformer decoder ($\textbf{Decoder}$) to estimate final poses $\textbf{P}_s$ as,
\begin{equation}
    \begin{aligned}
        & \mathbf{V} = \operatorname{\textbf{Conv1D}}(\mathbf{P}^\prime\textbf{W}_P) + \textbf{E}_{pos,0}\\
        & \textbf{P}_s = \operatorname{\textbf{Decoder}}(\mathbf{V}, (\textbf{S}_v;\textbf{S}_a; \mathbf{X})\textbf{W}_M)\textbf{W}_D,
     \end{aligned}
\end{equation}
where $\textbf{V}$ is our decoder's values embedded via a 1D convolutional layer, $\textbf{W}_P \in\mathbb{R}^{T/N\times T} $ is an interpolation matrix, and $\textbf{W}_M \in\mathbb{R}^{ (3KD+C) \times C}$ and $\textbf{W}_D\in\mathbb{R}^{C\times(K\cdot D)}$ are linear projection matrices.  

\subsection{Online Mutual Learning}

We further propose an online mutual learning that provides cross-supervision between refined input poses $\mathbf{P}^\prime$ and final predictions $\mathbf{P}_s$. HANet mutually optimize them by choosing online learning targets along their training losses. 

\textbf{Weighted Loss.} The objective of our loss function is minimizing the weighted $L_1$ norm between prediction and ground truth joint positions. Weighted loss tracks top-$k$ keypoints that are hard to predict based on the training loss by extending \cite{ohkm}. We refer to this weighted loss as $\mathcal{L}_{w}$, defined as follows.
\begin{equation}
    \mathcal{L}_w =\frac{1}{N_j} \sum_{j=1}^{N_j} v_{j} \|\mathbf{G}_j-\mathbf{P}_j\| + \frac{\lambda}{N_k} \sum_{k=1}^{N_k} v_{k} \|\mathbf{G}_k-\mathbf{P}_k\|,
\end{equation}
where $N_k,$ $\mathbf{G}_j$, $\mathbf{P}_j$, and $v_{j}$ denote the number of top-$k$ keypoints, ground truth, prediction, and visibility for joint $j$, respectively. This distance between prediction and ground truth is valid when a joint $j$ is visible. The first loss term penalizes all keypoint errors, while the second term only tracks the top-$k$ keypoint errors. 

\textbf{Online Loss.} Then, from the weighted loss $\mathcal{L}_{w}$, the online loss $\mathcal{L}_O$ is computed by,
\begin{equation}    
    \mathcal{L}_{O}^j=\begin{cases}  \mathcal{L}_{w}^{j}(\mathbf{P}_s, \mathbf{P}^\prime) & \mathcal{L}_{w}^{j}(\mathbf{G}, \mathbf{P}_s)<\mathcal{L}_{w}^{j}(\mathbf{G}, \mathbf{P}^\prime), \\ \mathcal{L}_{w}^{j}(\mathbf{P}^\prime, \mathbf{P}_s) & \text { otherwise, }\end{cases}
\end{equation}
where the first parameter of $\mathcal{L}_w$ is a target and the second one learns from the target. If $\mathcal{L}_{w}(\mathbf{G}, \mathbf{P}^\prime)$ is greater than $\mathcal{L}_{w}(\mathbf{G}, \mathbf{P}_s)$ for $j^{th}$ keypoint, we back propagate the refined input pose $\mathbf{P}^\prime$ to $\mathbf{P}_s$, and vice versa, final prediction $\mathbf{P}_s$ is penalized by $\mathbf{P}^\prime$. From these two components, we define our loss function as,
\begin{equation}
    \mathcal{L}=\mathcal{L}_{w}(\mathbf{G}, \mathbf{P}^\prime)+\lambda_s\mathcal{L}_{\boldsymbol{w}}(\mathbf{G}, \mathbf{P}_{\boldsymbol{s}})+\mathcal{L}_{O}(\mathbf{P}^\prime, \mathbf{P}_s),
\end{equation}
where $\lambda_s$ stands for weight of the final prediction error.

\begin{table*}[t]
\begin{center}
\setlength{\tabcolsep}{5.1pt}
\begin{tabular}{l|llllllll|l|l}
\toprule
\multirow{2}{*}{Method} & \multicolumn{8}{c|}{\textbf{PCK@0.2}} & \multicolumn{1}{c|}{\textbf{PCK@0.1}} & \multicolumn{1}{c}{\textbf{PCK@0.05}} \\  
& Head & Sho.          & Elb. & Wri. & Hip  & Knee & \multicolumn{1}{l|}{Ank.} & Avg. & Avg.                        & Avg.                         \\ \midrule
DKD \cite{dkd} & 98.3 & 96.6          & 90.4 & 87.1 & 99.1 & 96.0 & \multicolumn{1}{l|}{92.9} & 94.0 &                       -      &         -               \\
KFP (ResNet18) \cite{kfp}   & 94.7 & 96.3          & 95.2 & 90.2 & 96.4 & 95.5 & \multicolumn{1}{l|}{93.2} & 94.5 &                       -   &         -              \\ 
SimpleBaseline \cite{simplebaseline} & 97.5 & 97.8          & 91.1 & 86.0 & 99.6 & 96.8 & \multicolumn{1}{l|}{92.6} & 94.4 & 81.3                        & 56.9                         \\
SimpleBaseline + DeciWatch \cite{deciwatch} & 99.8 & 99.5          & \textbf{99.7} & \textbf{99.7} & 98.7 & 99.4 & \multicolumn{1}{l|}{96.5} & 98.8 & 94.1                        & 79.4                         \\ 
SimpleBaseline + \textbf{HANet}     & \textbf{99.9}  & \textbf{99.7} & \textbf{99.7} & \textbf{99.7} & \textbf{99.2} & \textbf{99.9} & \multicolumn{1}{l|}{\textbf{98.8}} & \textbf{99.6} & \textbf{98.3}                        & \textbf{91.9}  \\ \bottomrule    
\end{tabular}
\end{center}
\caption{Quantitative comparison on Sub-JHMDB dataset \cite{jhmdb} with prior works \cite{dkd,kfp,simplebaseline,deciwatch}.}
\label{table:JHMDB set}
\end{table*}

\begin{figure*}[t]
\centering
\includegraphics[width=1.0\textwidth]{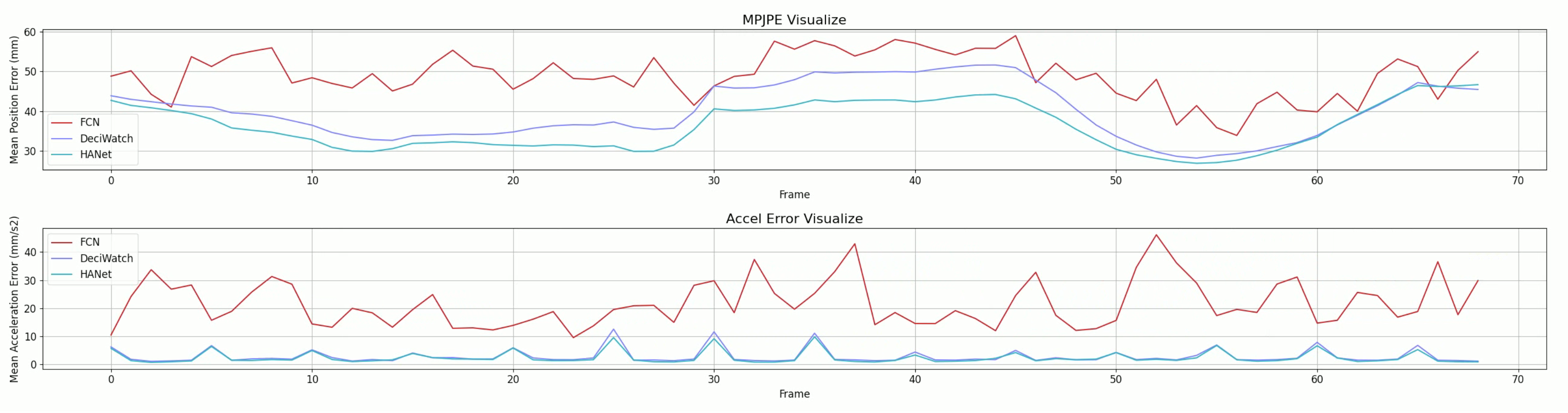}
\caption{\textit{MPJPE} and \textit{Accel} error comparison with prior works \cite{fcn,deciwatch} on a video from Human3.6M \cite{h36m} dataset. }
\label{fig:mpjpe error & Accel}
\end{figure*}

\begin{figure}[t]
\centering
\includegraphics[width=1.0\linewidth]{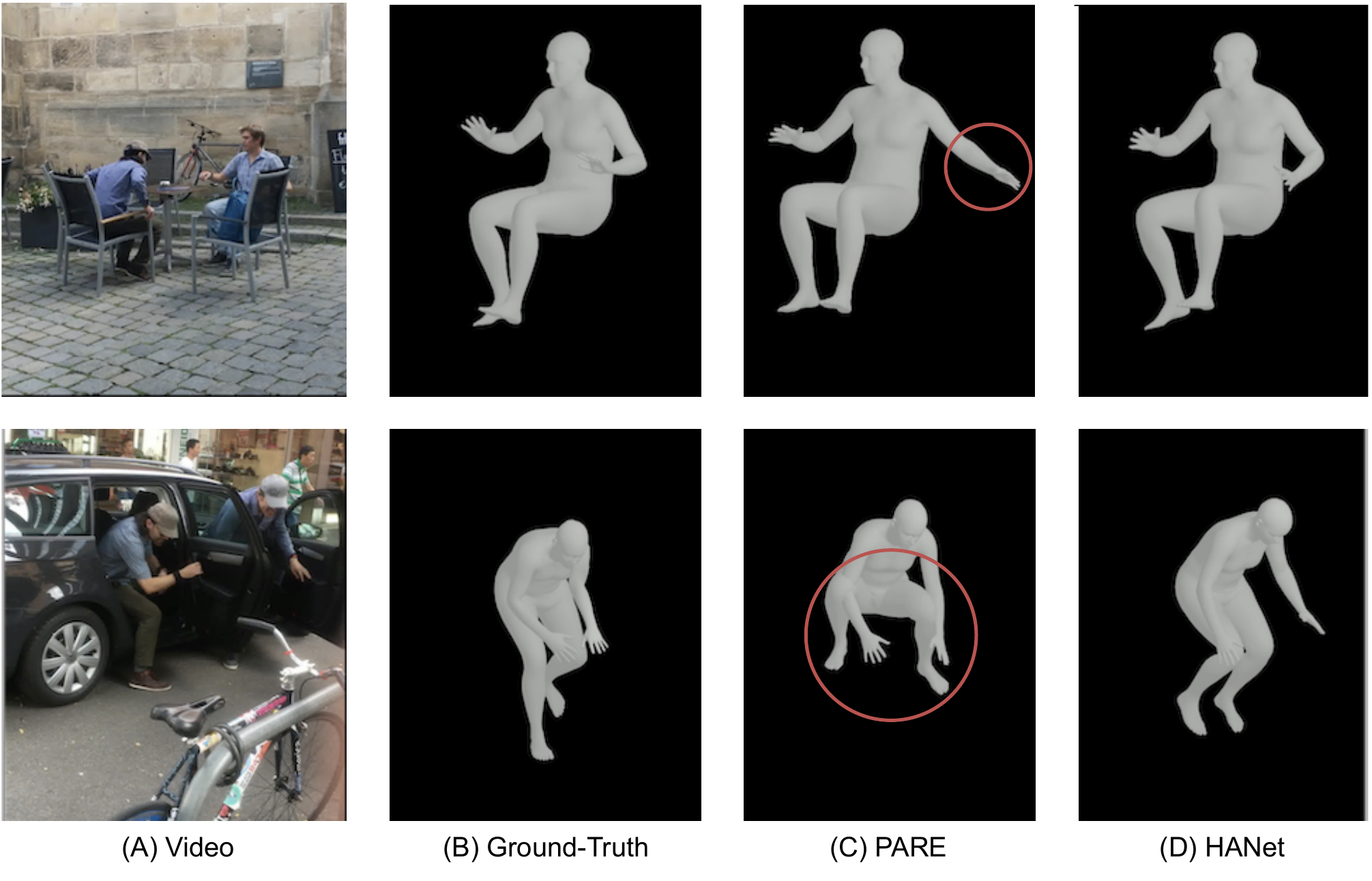}
\caption{Qualitative results of body mesh recovery on 3DPW \cite{3dpw}. We visualize that HANet alleviates occlusion and jitter of input poses well. We recommend to watch our supplementary video for jitter comparison.}
\label{fig:body recovery}
\end{figure}

\begin{table*}[t]
\begin{center}
\begin{tabular}{l|l|ll} 
 \toprule
Dataset                                         & Methods             & \textit{MPJPE} $\downarrow$           & \textit{Accel} $\downarrow$    \\ \midrule
\multirow{6}{*}{Human3.6M \cite{h36m}}          & FCN \cite{fcn}      & 54.6            & 19.2          \\
                                                & FCN + DeciWatch \cite{deciwatch} ($T$=101) & 52.8     & 1.5  \\
                                                & FCN + \textbf{HANet} ($T$=51) & \textbf{51.8}        & 2.0          \\
                                                & FCN + \textbf{HANet} ($T$=101) & 52.8        & \textbf{1.4}          \\
                                                & Mhformer \cite{li2022mhformer} & 38.3 & \textbf{0.8}\\
& Mhformer \cite{li2022mhformer} + \textbf{HANet} ($T$=101) & \textbf{35.4} 
& \textbf{0.8} 
\\\midrule
\multirow{4}{*}{3DPW \cite{3dpw}}               & PARE \cite{pare}      & 78.9            & 25.7          \\
                                                & PARE + DeciWatch \cite{deciwatch} ($T$=101) & 77.2     & 6.9  \\
                                                & PARE + \textbf{HANet} ($T$=51) & \textbf{74.6}        & 8.0           \\
                                                & PARE + \textbf{HANet} ($T$=101) & 77.1        & \textbf{6.8}           \\\midrule
\multirow{4}{*}{AIST++ \cite{aist}}             & SPIN \cite{spin} & 107.7              & 33.8          \\
                                                & SPIN + DeciWatch \cite{deciwatch} ($T$=101) & 71.3     & 5.7  \\
                                                & SPIN + \textbf{HANet} ($T$=51)  & \textbf{64.3}        & 6.4          \\
                                                & SPIN + \textbf{HANet} ($T$=101) & 69.2        & \textbf{5.4}           \\ \bottomrule
\end{tabular}
\end{center}
\caption{Quantitative comparison on the 3D pose estimation (Human3.6M \cite{h36m}) and body mesh recovery (3DPW \cite{3dpw}, AIST++ \cite{aist}).}
\label{table:3d}
\end{table*}

\begin{figure*}[t]
\centering
\includegraphics[width=1.0\linewidth]{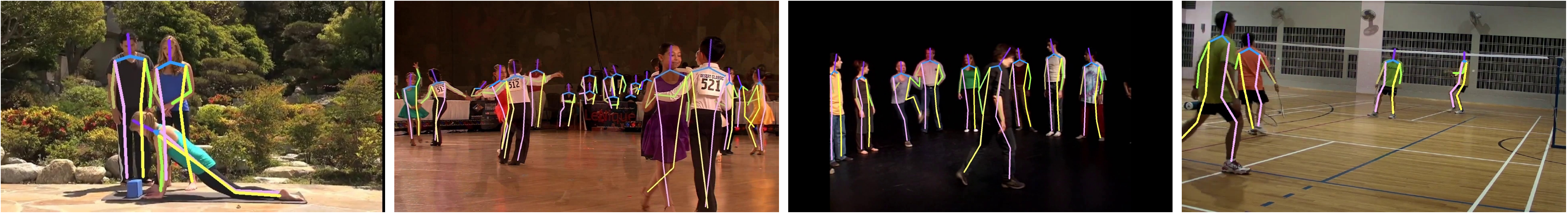}
\caption{Qualitative results of PoseTrack 2018 test set. We demonstrate that HANet performs well at multi-human 2D pose estimation, even in occlusion, motion blur, and crowd environments.
}
\label{fig:complex scene}
\end{figure*}

\setlength{\tabcolsep}{5.9pt}
\begin{table*}[]
\begin{center}
\begin{tabular}{l|l|lllllll|l}
\toprule
Dataset & Method                        & Head & Shoulder & Elbow & Wrist & Hip & Knee & Ankle & Mean\\ \midrule
\multirow{4}{*}{PoseTrack17 Validation \cite{posetrack2017}}  
& HRNet \cite{hrnet}        & 82.1 & 83.6 & 80.4 & 73.3 & 75.5 & 75.3 & 68.5 & 77.3 \\
& PoseWarper \cite{posewarper}& 81.4 & 88.3 & 83.9 & 78.0 & 82.4 & 80.5 & 73.6 & 81.2 \\
& DCPose \cite{dcpose}      & 88.0 & 88.7 & 84.1 & 78.4 & 83.0 & 81.4 & 74.2 & 82.8 \\
& DCPose + \textbf{HANet}   & \textbf{90.0} & \textbf{90.0} & \textbf{85.0} & \textbf{78.8} & \textbf{83.1} & \textbf{82.1} & \textbf{77.1} & \textbf{84.2} \\ \midrule
\multirow{3}{*}{PoseTrack18 Validation \cite{posetrack2018}} 
& PoseWarper \cite{posewarper}  & 79.9 & 86.3   & 82.4  & 77.5  & 79.8 & 78.8 & 73.2  & 79.7 \\
& DCPose \cite{dcpose}          & 84.0 & 86.6     & 82.7  & 78.0  & \textbf{80.4} & 79.3 & 73.8  & 80.9 \\ 
& DCPose + \textbf{HANet} & \textbf{86.1} & \textbf{88.5} & \textbf{84.1} & \textbf{78.7} & 79.0 & \textbf{80.3} & \textbf{77.4} & \textbf{82.3}\\\bottomrule
\end{tabular}
\end{center}
\caption{Quantitative comparison on the validation sets of PoseTrack2017 \cite{posetrack2017}  and PoseTrack2018 \cite{posetrack2018}.}
\label{table:2017 validation and test set}
\end{table*}

\section{Experiments}

In this section, we discuss our extensive experiments and demonstrate that our proposed method refines input poses well and can generally be applied to 2D pose estimation, 3D pose estimation, body mesh recovery, and sparsely-annotated 2D pose estimation tasks.

\subsection{Datasets}

We evaluate our framework on four tasks and report experimental results on various benchmark datasets. First, we use the dataset Sub-JHMDB \cite{jhmdb} for 2D pose estimation. Second, we validate HANet on PoseTrack2017 \cite{posetrack2017} and PoseTrack2018 \cite{posetrack2018} for sparsely-annotated multi-human 2D pose estimation. For 3D pose estimation, we select the most generally used dataset Human3.6M \cite{h36m}. Lastly, we verify our model on an in-the-wild dataset 3DPW \cite{3dpw}, and a dance dataset AIST++ \cite{aist} with fast and diverse actions, for body mesh recovery. 

\subsection{Estimators}

We trained our model using the estimated 2D coordinates, 3D coordinates, or SMPL parameters as input. Specifically, we use off-the-shelf estimators such as SimpleBaseline \cite{simplebaseline} for Sub-JHMDB, DCPose \cite{dcpose} for PoseTrack, FCN \cite{fcn} and Mhformer \cite{li2022mhformer} for Human3.6M, PARE \cite{pare} for 3DPW, and SPIN \cite{spin} for AIST++.

\subsection{Evaluation Metrics}
For 2D pose estimation, we adopt the Percentage of Correct Keypoints (\textbf{PCK}) following \cite{dkd, kfp, deciwatch}, where the matching thresholds are set as 20\%, 10\%, and 5\% of the bounding box size under pixel level, and mean Average Precision (\textbf{mAP}) only for visible joints following \cite{posetrack2017,poseflow,simplebaseline,hrnet}. For 3D pose estimation and body mesh recovery, we adopt Mean Per Joint Position Error (\textit{MPJPE}) and the mean Acceleration error (\textit{Accel}) following \cite{fcn,spin,pare}.

\subsection{Implementation Details}

In our experiments, we consider the previous and next velocity of keypoints, $\boldsymbol{v}_p$ and $\boldsymbol{v}_n$. We set a different length $T$ of the sliding window and an interval of frames $N$ for each task. For sparsely-annotated multi-human 2D pose estimation, a long sliding window may not be useful because multi-human 2D pose estimation often involves more severe occlusion and motion blur than single-human pose estimation. So, we use $T = 5$ and $N=1$ and fix the input image size to $384 \times 288$. Furthermore, PoseTrack datasets \cite{posetrack2017,posetrack2018} are sparsely annotated, we enlarge the bounding box by 25\% from the annotated frame and use it to crop the supplement frames. Within a sliding window, we only trained with annotated frames. We use an AdamW optimizer \cite{adamw}, set an initial learning rate as $1 \times 10^{-3}$, warm it up at the first 5 epochs, and then decay in a cosine annealing manner. For weighted loss $\mathcal{L}_w$, we set $\lambda$ as 0.5 to match the normalization degree of the first term because we set $N_k$ as half of the number of joints. 

\subsection{Comparisons}

\textbf{2D Pose Estimation.} As illustrated in Fig.~\ref{fig:jhmdb qualitative}, we visualize the estimated pose from \cite{simplebaseline}, existing  state-of-the-art method \cite{deciwatch}, and our method for Sub-JHMDB \cite{jhmdb}, which is a 2D single-human pose dataset. The input pose estimator SimpleBaseline \cite{simplebaseline} showed a performance of 65.00 on an intricate metric of PCK@0.05, and DeciWatch \cite{deciwatch} showed a result of 72.22. Above all, we achieve the remarkable results of 96.67. In addition, we quantitatively validate our model against current state-of-the-art models \cite{dkd,kfp,simplebaseline,deciwatch}. Although the performance for PCK@0.2 may look similar, Table~\ref{table:JHMDB set} shows that the performance improvement of our model is noticeable as the threshold reduces from PCK@0.2 to PCK@0.05.

\textbf{3D Pose Estimation \& Body Mesh Recovery.} In addition, we demonstrate that our method can be applied to different tasks such as 3D pose estimation and body mesh recovery. We input 3d pose coordinates or SMPL parameters estimated from \cite{pare,spin,fcn,li2022mhformer} and refine their results. First, we validate HANet compared to state-of-the-art methods \cite{pare} for body mesh recovery on 3DPW \cite{3dpw} dataset. We visualize results of body mesh recovery for highly-occluded scenes. As illustrated in Fig.~\ref{fig:body recovery}, \cite{pare} shows large positional difference. However, our model shows accurate estimation results because our model learns sufficient temporal information through keypoint kinematic features. We further visualize \textit{MPJPE} and \textit{Accel} error with FCN \cite{fcn} and DeciWatch \cite{deciwatch} on Human3.6M \cite{h36m} dataset. In Fig.~\ref{fig:mpjpe error & Accel}, we show that our framework reduces \textit{MPJPE} more while maintaining a lower \textit{Accel} error. 

We also validate our framework and present the quantitative results on Table~\ref{table:3d} for 3D pose estimation and body mesh recovery. We compare the \textit{MPJPE} and \textit{Accel} with input pose estimators \cite{fcn,li2022mhformer,spin,pare} and DeciWatch \cite{deciwatch}, which efficiently improved performance by sampling keyframes taking advantage of continuity of human motions. We report two version of our HANet that uses a different sliding window length $T = 51$ and $T = 101$. When $T=101$, i.e., learn more temporal relationships between consecutive poses, we observe that \textit{Accel} metric is improved more. Conversely, when we set $T=51$, we demonstrate that our method reduces the need for a long sliding window and efficiently improved \textit{MPJPE} and \textit{Accel} errors.

\textbf{Sparsely Annotated Pose Estimation.} We also conduct experiments on PoseTrack2017 \cite{posetrack2017} and PoseTrack2018 \cite{posetrack2018}, which is sparsely-annotated datasets. In Table~\ref{table:2017 validation and test set}, we compare our model with state-of-the-art methods \cite{hrnet,posewarper,dcpose} and the input pose estimator \cite{dcpose} on the validation set of PoseTrack2017 and PoseTrack2018. We achieve state-of-the-art results and demonstrate that our method improves performance (\textbf{mAP}) on sparsely-annotated multi-human 2D pose estimation. In addition, we visualize that our method shows remarkable results in occlusion, crowded scenes, fast motion, and challenging pose, as shown in Fig.~\ref{fig:complex scene}.  

\subsection{Ablation Study}

We validate the effect of our component and keypoint kinematic features with an extensive ablation study. We perform the experiments on Sub-JHMDB \cite{jhmdb} and compare results in Table~\ref{table:Ablation study on component} - Table~\ref{table:Ablation Study on kinematic features}. We use  \textbf{PCK} for the evaluation metric.

\textbf{Hierarchical Encoder.} First, we compare and report the effect of our hierarchical encoder at rows 1-2 and scale of the encoder at rows 5-7 in Table~\ref{table:Ablation study on component}. When we replace the hierarchical encoder to normal transformer encoder, performance significantly drops (2.3\%) for PCK@0.05. In addition, we scale the number of encoder and decoder layers $N_L$ to $4$ and grew one by one. It gradually improves performance up to 5 layers (4.7\%) and decreases (1.7\%). 

\setlength{\tabcolsep}{3.8pt}
\begin{table}[t]
\begin{center}
\begin{tabular}{cc|cc|c|c|c}
\toprule
\multicolumn{4}{c|}{\textbf{Component}} & \multirow{3}{*}{\textbf{PCK@0.2}}       & \multirow{3}{*}{\textbf{PCK@0.1}}       & \multirow{3}{*}{\textbf{PCK@0.05}}\\ 
\multirow{2}{*}{$N_L$} & \multirow{2}{*}{HE}        & \multicolumn{2}{|c|}{OML} & & &\\

& & $L_1$ & $L_2$ & & & \\  \midrule
4 & & & & 98.2 & 95.9 & 83.8 \\
4 & \checkmark & & & 99.2 & 97.0 & 86.1 \\ 
4 & \checkmark & & \checkmark & 99.1 & 96.3 & 85.8 \\
4 & \checkmark & \checkmark & & 99.3 & 97.2 & 86.7 \\ 
4 & \checkmark & \checkmark & \checkmark & 99.4 & 97.3 & 87.2 \\
5 & \checkmark & \checkmark & \checkmark & \textbf{99.6} & \textbf{98.3} & \textbf{91.9} \\  
6 & \checkmark & \checkmark & \checkmark & 99.5 & 97.7 & 90.2 \\ \bottomrule
\end{tabular}
\end{center}
\caption{Ablation study to observe the contribution of each component of our model on JHMDB \cite{jhmdb}.}
\label{table:Ablation study on component}
\end{table}

\textbf{Online Mutual Learning.} We verify the effectiveness of our proposed online mutual learning (OML) at rows 2-5. Rows 2 and 4 indicates that OML with $L_1$ norm has a significant impact on performance improvement (0.6\%) for PCK@0.05. We also conduct experiments for OML with $L_2$ norm (row 3 and 5) and found that OML has improved performance by jointly using the $L_1$ norm and $L_2$ norm. We proceed with the rest of the ablation study using the row 6 version that showed the best performance.

\textbf{Keypoint Kinematic Features.} We also conduct ablation studies on keypoint kinematic features on Table~\ref{table:Ablation Study on kinematic features}. Here, except for the last row, the encoder does not take concatenated previous frame and subsequent frame features. The first row is the minimum version of HANet that removes all features associated with keypoint kinematic features, and the last row is the complete version of HANet considering all keypoint kinematic features: flows, velocity, acceleration, and WB (weight and bias). From the top, the rows in turn indicate the addition of flow, velocity, acceleration, and WB. We could observe significant performance impact of each keypoint kinematic features in the order of velocity (2.2\%), flow (1.9\%), acceleration (0.9\%), and WB (0.4\%) for PCK@0.05.

\setlength{\tabcolsep}{1.8pt}
\begin{table}[t]
\begin{center}
\begin{tabular}{cccc|ccc}
\toprule
\multicolumn{4}{c|}{\textbf{Component}}        &\multirow{2}{*}{\textbf{PCK@0.2}} &\multirow{2}{*}{\textbf{PCK@0.1}} &\multirow{2}{*}{\textbf{PCK@0.05}}\\ 
Flow  & Vel. & Accel.            & WB   & & & \\ \midrule 
           &      &              &      & 99.1          & 96.3          & 86.5  \\ 
\checkmark &      &              &      & 99.4          & 97.4          & 88.4  \\
\checkmark & \checkmark &        &      & 99.4          & 97.6          & 90.6  \\ 
\checkmark & \checkmark & \checkmark &  & 99.5          & 98.1          & 91.5  \\
\checkmark & \checkmark & \checkmark & \checkmark& \textbf{99.6} & \textbf{98.3} & \textbf{91.9}\\ \bottomrule
\end{tabular}
\end{center}
\caption{Ablation study of keypoint kinematic features on JHMDB. WB stands for weight and bias of velocity and acceleration.}
\label{table:Ablation Study on kinematic features}
\end{table}
\section{Conclusion}

In this work, we construct a hierarchical transformer encoder and exploit the keypoint kinematic features to address occlusion and mitigate jitter. We demonstrate that explicitly leveraging flow, velocity, and acceleration can capture the keypoint's temporal dependencies better. Furthermore, HANet effectively addresses the occlusion issue by learning spatio-temporal relationships between consecutive frames' poses through multi-scale feature maps and achieve state-of-the-art performance on a variety of pose estimation tasks. We are interested in extending our model to an end-to-end model, which can be combined with our approach to produce more accurate results.

\textbf{Acknowledgement.} This work was partially supported by Institute of Information \& communications Technology Planning \& Evaluation (IITP) grant funded by the Korea government(MSIT) (No. 2019-0-00079, Artificial Intelligence Graduate School Program (Korea University), No. 2022-0-00984, Development of Artificial Intelligence Technology for Personalized Plug-and-Play Explanation and Verification of Explanation).

{\small
\bibliographystyle{ieee_fullname}
\bibliography{egbib}
}

\end{document}